\documentclass[runningheads]{llncs}
\usepackage{graphicx}
\usepackage{amsmath,amssymb} 
\usepackage{mathtools}
\usepackage{dsfont}
\usepackage{color}
\usepackage{bm}
\usepackage{booktabs}
\usepackage{longtable}
\usepackage{colortbl}
\usepackage{tabulary,multirow,overpic,xcolor}
\usepackage[caption=false]{subfig}
\usepackage[pagebackref=false,breaklinks=true,letterpaper=true,
colorlinks, urlcolor=blue, citecolor=blue]{hyperref}
\newlength\savewidth

\usepackage{tablefootnote}
\definecolor{orcidlogocol}{HTML}{A6CE39}
\definecolor{scholar}{HTML}{2DA9FA}

\begin{document}
	
\title{Neural Network Encapsulation} 

\titlerunning{Neural Network Encapsulation}

\author{Hongyang 
	Li\inst{1}\thanks{Contact email: \texttt{yangli@ee.cuhk.edu.hk}}
	~~Xiaoyang Guo\inst{1}
	~~Bo Dai\inst{1}
	\\Wanli Ouyang\inst{2}~~~Xiaogang Wang\inst{1}
}

\authorrunning{H. Li \textit{et al.}}
%

\institute{The Chinese University of Hong Kong \and
The University of Sydney
}


%
\maketitle              
\begin{abstract}
A capsule is a collection of neurons which represents different variants of a pattern in the network. 
The routing scheme ensures only certain  capsules which resemble lower counterparts in the higher layer should be activated.
However, the computational complexity becomes a bottleneck for scaling up  to larger networks, 
as lower capsules need to correspond to each and every higher capsule. 
To resolve this limitation, we approximate the routing process with two branches: 
a master branch which collects primary information from its direct contact in the lower layer 
and an aide branch that replenishes master based on pattern variants encoded in other lower capsules. 
Compared with previous iterative and unsupervised routing scheme, these two branches are 
communicated in a fast, supervised and one-time pass fashion.
The complexity and runtime of the model are therefore decreased by a large margin.
Motivated by the routing to make higher capsule have agreement with lower capsule,
we extend the mechanism as a compensation for the rapid loss of information in nearby layers.
We devise a feedback agreement unit to send back higher capsules as feedback. It could be regarded as an additional regularization to the network. The feedback agreement is achieved by comparing the optimal transport divergence between two distributions (lower and higher capsules). Such an add-on witnesses a unanimous gain in both capsule and vanilla networks. Our proposed EncapNet performs favorably better against previous state-of-the-arts on CIFAR10/100, SVHN and a subset of ImageNet.

\end{abstract}

\section{Introduction}

Convolutional neural networks (CNNs) \cite{mnist} have been proved to be quite successful in modern deep learning architectures \cite{alexnet,vgg,googlenet,resNet} and achieved better performance in various computer vision tasks \cite{li2017zoom,li2017we,czz_tracking}. By tying the kernel weights in convolution, CNNs have the translation invariance property that can identify the same pattern irrespective of the spatial location.
Each neuron in CNNs is a scalar  and can detect different (low-level details or high-level regional semantics) patterns layer by layer. However, in order to detect the same pattern with various variants in viewpoint, rotation, shape, \textit{etc.}, we need to stack more layers, which tends to ``memorize the dataset rather than generalize a solution'' \cite{cap_em_blog}.   
%
%

A capsule \cite{capsule,cap_EM} is a group of neurons whose output, in form of a vector instead of a scalar, represents various perspectives of an entity, such as pose, deformation, velocity, texture, object parts or regions, \textit{etc}. It captures the existence of a feature \textit{and} its variant. Not only does a capsule detect a pattern but also it is trained to learn the many variants of the pattern. This is what CNNs are incapable of.
The concept of capsule provides a new perspective
on feature learning via instance parameterization of entities (known as capsules) to encode different variants 
within a capsule structure, thus achieving the feature equivariance property\footnote{Equivariance is the detection of feature patterns that can transform to each other.} and being robust to adversaries. 
Intuitively, the  capsule  detects a pattern (say a face) with a certain variant (it rotates 20 degree clockwise) rather than realizes that the pattern matches a variant in the higher layer.

\begin{figure}[t]
	\begin{center}
		\includegraphics[width=0.95\textwidth]{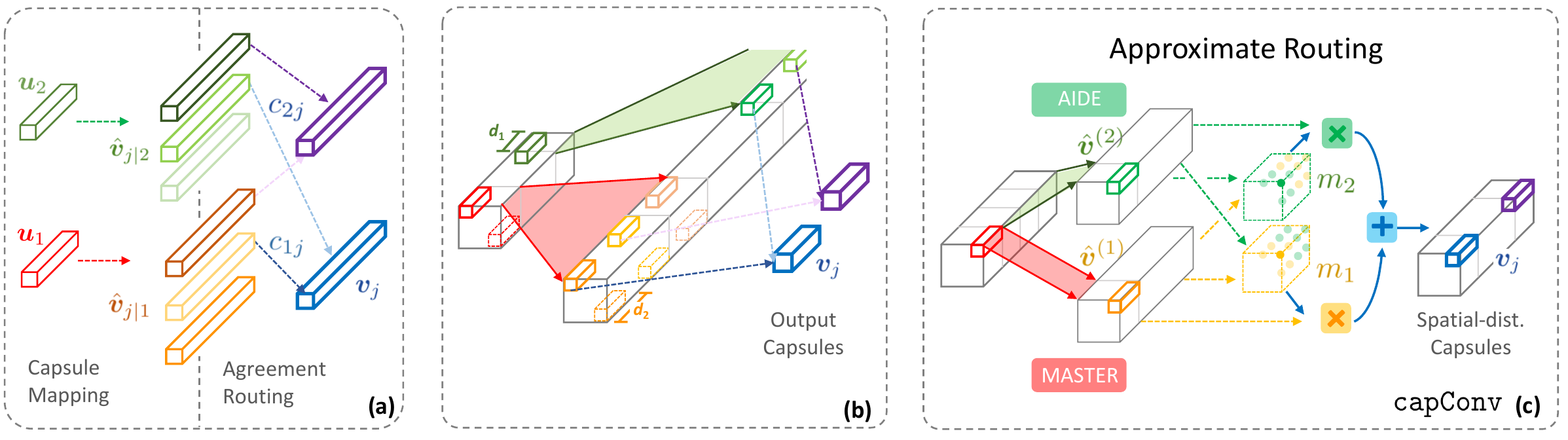}
	\end{center}
	\vspace{-.5cm}
	\caption{
		(a) One capsule operation includes a capsule mapping and an agreement routing. (b) Capsule implemented in a convolutional manner by \cite{capsule,cap_EM} where  lower capsules are mapped into the space of \textit{all} higher capsules and then routed to generate the output capsule. (c) Our proposed \texttt{capConv} layer: approximate routing with master and aide interaction to ease the computation burden in the current design in (b). 
	}
	\label{fig:capsule_block}
	\vspace{-.4cm}
\end{figure}

One basic capsule layer consists of two steps:  \textit{capsule mapping} and   \textit{agreement routing}, which is depicted in Fig. \ref{fig:capsule_block}(a). The input capsules are first mapped into the space of their higher counterparts via a transform matrix. 
Then the routing process involves all capsules between adjacent layers to communicate by the routing co-efficients; 
it ensures only certain lower capsules which resemble higher ones (in terms of cosine similarity) can pass on information and activate the higher counterparts.
Such a scheme can be seen as a feature clustering  and is optimized 
by coordinate descent through several iterations. 
%
However, the computational complexity in the first mapping step is the main bottleneck to apply the capsule idea in CNNs; lower capsules have to generate correspondence for every higher capsule (\textit{e.g.}, a typical choice \cite{capsule} is 2048 capsules with 16 dimension, resulting in 8 million parameters in the transform matrix). 

To tackle this drawback, we propose an  alternative to estimate the original routing summation by introducing  two branches: one is the \texttt{master} branch that serves as the primary source from the direct contact capsule in the lower layer; another is the \texttt{aide} branch that strives for searching other pattern variants along the channel and replenishes side information to \texttt{master}. These two branches are intertwined by their co-efficients so that feature patterns encoded in lower capsules could be fully leveraged and exchanged. Such a one-pass approximation is fast, light-weight and supervised, compared to the current iterative, short-lived and unsupervised routing scheme.

Furthermore, the routing effect in making higher capsule have agreement with lower capsule can be extended as a direct loss function. In deep neural networks, 
information is inevitably lost through stack of layers. To reduce the rapid loss of information in nearby layers, a loss function can be included to enforce that neurons or capsules in the higher layer can be used for reconstructing the counterparts in lower layers.
Based on this motivation, we devise an agreement feedback unit which sends back higher capsules as a feedback signal to better supervise feature learning. This could be deemed as a regularization on network. Such a feedback agreement  is achieved by measuring the distance between the two distributions using optimal transport (OT) divergence, namely the Sinkhorn loss. The OT metric (\textit{e.g.}, Wasserstein loss) is promised to be superior than other options to modeling data on general space. 
This add-on regularization is inserted during training and disposed of for inference. The agreement enforcement has witnessed a unanimous gain in both  capsule  and vanilla neural networks.

Altogether, bundled with the two mechanisms aforementioned,  we {(i)} encapsulate the neural network in an approximate routing scheme with master/aide interaction, {(ii)} enforce the network's regularization by an agreement feedback unit via optimal transport divergence. The proposed capsule network is denoted as EncapNet and performs superior against previous state-of-the-arts for image recognition tasks on CIFAR10/100, SVHN and a subset of ImageNet. The code and dataset are available \href{https://github.com/hli2020/nn_capsulation}{\texttt{https://github.com/hli2020/nn\_capsulation}}.

\vspace{-.2cm}
\section{CapNet: Agreement Routing Analysis}
\subsection{Preliminary: capsule formulation}
Let $\bm{u}_i, \bm{v}_j$ denote the input and output capsules in a layer, where $i, j$ indicates the index of capsules. 
The  dimension and the number of capsules at input and output are $d_1, d_2, n_1, n_2$, respectively, \textit{i.e.}, 
$\{ \bm{u}_i  \in \mathds{R}^{d_1}  \}_{i=1}^{n_1}, \{ \bm{v}_j  \in \mathds{R}^{d_2}  \}_{j=1}^{n_2}$. The first step 
is a mapping from lower capsules to higher counterparts:
$	\hat{\bm{v}}_{j | i} = \bm{w}_{ij} \cdot \bm{u}_i, $
where $\bm{w}_{ij} \in \mathds{R}^{d_1 \times d_2}  $ is a transform matrix
and we define  the intermediate output $\hat{\bm{v}}_{j | i} \in \mathds{R}^{d_2}$ as \textit{mapped activation} (called prediction vector in \cite{capsule}) from $i$ to $j$.
The second step is an agreement routing process to 
aggregate all lower capsules into higher ones.
%
The mapped activation is multiplied by a routing coefficient $c_{ij}$ 
through several iterations in an unsupervised  manner:
%
$\bm{s}_j^{(r)}  = \sum_i 
c_{ij}^{(r)}  \hat{\bm{v}}_{j | i}$.

This is where the highlight of capsule idea resides in. It could be deemed as a voting process: the activation of higher capsules should be entirely dependent on the resemblance from the  lower entities. 
Prevalent routing algorithms include the coordinate descent optimization \cite{capsule} and the Gaussian mixture clustering via Expectation-Maximum (EM)  \cite{cap_EM}, to which we refer as \texttt{dynamic} and \texttt{EM} routing, respectively.
For \texttt{dynamic} routing, given
$
b_{ij}^{(0)} \leftarrow 0, 
r \leftarrow 0$, we have:
\begin{gather}
{b}_{ij}^{(r+1)} \leftarrow {b}_{ij}^{(r)} +   \hat{\bm{v}}_{j | i}  \cdot \bm{v}_j^{(r)},
\end{gather}
%
where $b$ is the softmax input to obtain 
$c$; $\bm{v}^{(r)}$ is computed from $\bm{s}^{(r)}$ via  $\texttt{squash}(\cdot)$, \textit{i.e.}, $\bm{v} = \frac{ \| \bm{s} \|^2}{ 1 + \| \bm{s}\|^2}  \frac{\bm{s}}{\| \bm{s}\|}$.
The update of the routing co-efficient is conducted in a coordinate descent manner which optimizes $c$ and $\bm{v}$ alternatively.
For \texttt{EM} routing, given $c_{ij}^{(0)} \leftarrow 1/n_2, r \leftarrow 0$, and the activation response of input capsules $a_i$, we iteratively aggregate input capsules into $d_2$ Gaussian clusters:
\begin{gather}
a_j^{(r)}, \bm{\mu}_j^{(r)}, \bm{\sigma}_j^{(r)} \leftarrow \texttt{M-step} \big [ a_i, c_{ij}^{(r)}, \hat{\bm{v}}_{j | i} \big ], 
\\ 
c_{ij}^{(r+1)} \leftarrow 
\texttt{E-step} \big[	a_j^{(r)},  p_{j | i}
\big( \hat{\bm{v}}_{j | i}, \bm{\mu}_j^{(r)}, \bm{\sigma}_j^{(r)} \big)
\big],
\end{gather}
where the  mean of  cluster $\bm{\mu}_j$ is deemed as the output capsule $\bm{v}_j$. 
\texttt{M-step} generates the activation $a_j$  alongside the mean and std w.r.t. higher capsules; these variables are further fed into  \texttt{E-step} to update the routing co-efficients $c_{ij}$.
The output from a capsule layer is thereby obtained after iterating $R$ times. 
%
%

\subsection{Agreement routing analysis in CapNet}\label{sec:routing-by-agreement-in-capnet} 
\begin{figure}[t]
	\centering
	\includegraphics[width=0.98\textwidth]{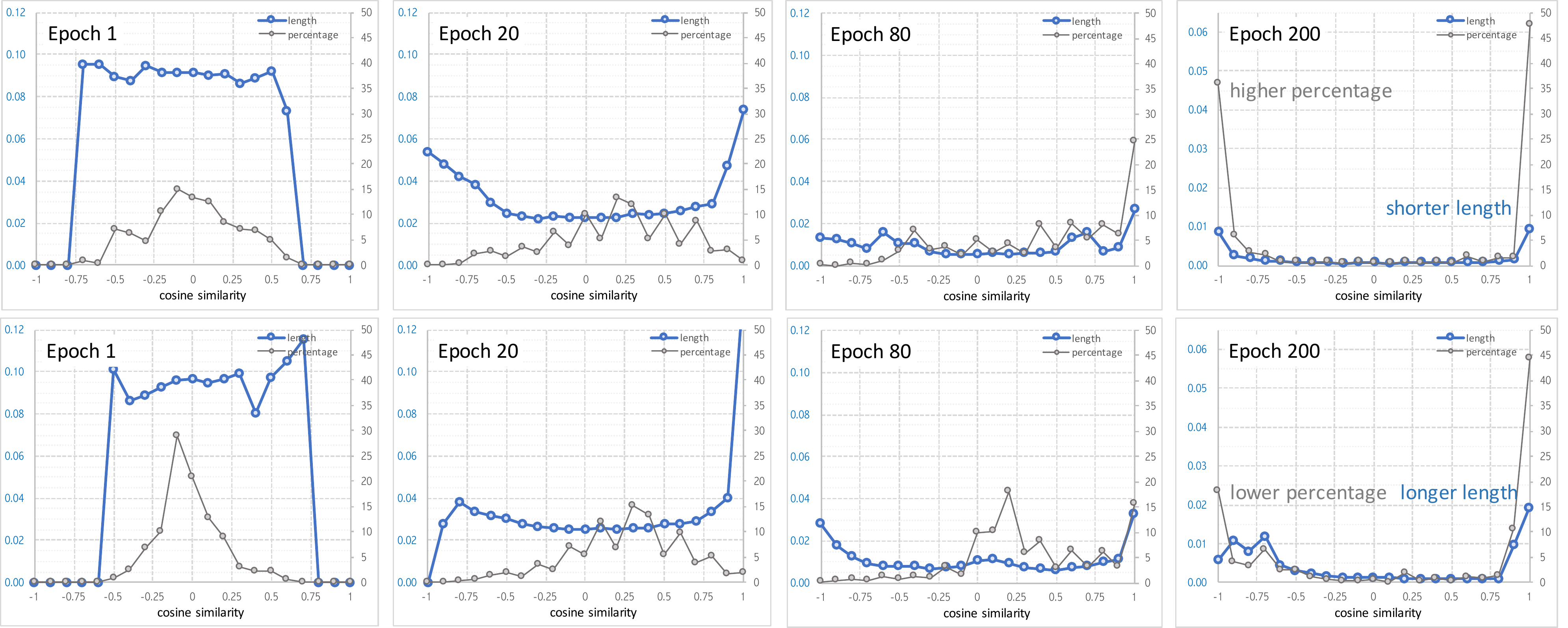}
	\vspace{-.2cm}
	\caption{
		Training dynamics  as network evolves. Routing tends to magnify and pass on pattern variants of lower capsules to higher ones which mostly resemble the lower counterparts. 
		\textbf{Top}: \texttt{Dynamic} routing. \textbf{Bottom}: \texttt{EM} routing. We show the cosine similarity between $\bm{v}_j$ and the mapped lower capsules, \textit{i.e.}, \texttt{cos\_sim}($ \bm{v}_j, \hat{\bm{v}}_{j|i}$). 
		Blue line represents the average (across all  samples) length $ \| \hat{\bm{v}}_{j|i} \|$ and gray indicates the percentage (\%) of how many lower capsules $i$'s agree with  $j$ at a given resemblance. 
	}
	\label{fig:routing_process}
	\vspace{-.5cm}
\end{figure}

\textbf{Effectiveness of the agreement routing.} 
Figure \ref{fig:routing_process} illustrates the training dynamics on routing between adjacent capsules as the network evolves.
In essence, the routing process is a weighted average from all lower capsules to the higher entity (Eqn.(\ref{elementwise})). Intuitively, given a sample which belongs to the $j$-th class, the network tries to optimize capsule learning such that the length (existence probability) of $\bm{v}_j$  in the final capsule layer should be the largest. This requires the magnitude of its lower counterparts who resemble capsule $j$ should occupy a majority and have a higher length compared to others that are dissimilar to $j$.
Take the top row of \texttt{Dynamic} case for instance. At the first epoch, the kernel weights $\bm{w}_{ij}$ are initialized with Gaussian hence most capsules are orthogonal to each other and have the same length. As training goes (epoch 20 and 80), the percentage and length of ``blurring'' capsules, whose cosine similarity is around zero, goes down and the distribution evolves into a polarization: the most similar and  dissimilar capsules gradually take the majority and hold a higher length than other $i$'s. As training approaches termination (epoch 200), such a phenomenon is further polarized and the network is at a stable state where the most resembled and non-resembled capsules have a higher percentage and length than the others. 
The role of agreement routing is to adjust the magnitude and relevance from lower capsules to higher capsules, such that the activation of relevant higher counterparts could be appropriately turned on and the pattern information from lower capsules be passed on.

The analysis for \texttt{EM} routing draws a unanimous conclusion. The polarization phenomenon is further intensified (\textit{c.f.} (h) vs (d) in Fig. (\ref{fig:routing_process})). The percentage of dissimilar capsules is lower (20\% vs 37\%) whilst the length of similar capsules is higher (0.02 vs 0.01): implying that \texttt{EM} is potentially a better routing solution than \texttt{dynamic}, which is also verified by (a) vs (b) in Table \ref{net_compare}.

Moreover, it is observed that replacing scalar neurons in traditional CNNs with vector capsules and routing is effective, \textit{c.f.} (a-b) vs (c) in Table \ref{net_compare}. We adopt the same blob shape for each layer in vanilla CNNs for fair comparison. However, when we increase the parameters of  CNNs  to the same amount as that of CapNet, the former performs better in  (d). Due to the inherent design, CapNet requires  more parameters than the traditional CNNs, \textit{c.f.} (a) vs (c) in Table \ref{net_compare} with around 152 Mb for CapNet vs 24 Mb for vanilla CNNs.  

The capsule network is implemented in a  group convolution fashion by \cite{capsule,cap_EM}, which is depicted in Fig. \ref{fig:capsule_block}(b). It is assumed that  the vector capsules are placed in the same way as the scalar neurons in vanilla CNNs. The \textit{spatial} capsules in a  channel share the same transform kernel 
since they search for the same patterns in different locations. The \textit{channel} capsules own different kernels as they represent various patterns encapsulated in a group of neurons.

\textbf{Computational complexity in CapNet.} 
From an engineering perspective,
the original design for capsules in CNN structure (see Fig. \ref{fig:capsule_block}(b))  is to save computation cost in the capsule mapping step; otherwise it would take $64\times$ more kernel parameters (assuming spatial size is 8) to fulfill the mapping step. However, the burden is not eased effectively since step one has to generate a mapping for  {\textit{each and every}} capsule $j$ in the subsequent layer. The  output channel size of the transform kernel in Tab. \ref{net_compare}(a-b) is 1,048,576 ($16\times 32\times 2048$). 
If we feed the network with a batch size of 128 (even smaller option, \textit{e.g.}, 32), OOM (out-of-memory) occurs due to the super-huge volume of the transform kernel.
The subtle difference of parameter size between \texttt{dynamic} and \texttt{EM} is that additionally the latter has a larger convolutional output before the first capsule operation to generate activations; and it has a set of trainable parameters in the \texttt{EM} routing.
Another impact to consider is  the routing co-efficient matrix of size $n_1 \times n_2$, the computation cost for this part is lightweight though and yet it takes longer runtime than traditional CNNs due to the routing iteration times $R$ to update $c$, especially for \texttt{EM} method that involves two update alternations.

Inspired by the routing-by-agreement scheme to aggregate  feature patterns  in the network and bearing in mind that the current  solution has a large computation complexity, we resort to some alternative that both inherits the spirit of routing-to-interact among capsules and implements such a process in a fast and accurate fashion. This leads to the proposed scheme stated below.

\renewcommand\arraystretch{1.2}
\setlength{\tabcolsep}{3pt}
\begin{table}[t]
	\centering
	\caption{Comparison of vanilla CNN, CapNet \cite{capsule,cap_EM} and  EncapNet. All models have a depth of  six layers and are compared
		via  (i) the number of  model parameters (Mb), (ii) memory consumption (MB, at a given batch size), (iii) runtime (second per  batch size) and (iv) performance (error rate \%). 
		8 and 4 is the largest batch size that can fit in 
		memory\protect 
		\footnotemark [4]   
		for \texttt{dynamic} and \texttt{EM} routing.  Metric (ii) and (iii) are measured on CIFAR-10.
	}\label{net_compare}
	\vspace{-.1cm}
	\scriptsize	{
		\begin{tabular}{l || l  || l |  c ||  c  ||  c}
			\hline
			\multirow{1}{*}{method} &  \multirow{1}{*}{param \#} & \multirow{1}{*}{mem. size} & runtime   & CIFAR-10 & MNIST \\
			\hline
			(a) CapNet, \texttt{dynamic} &  151.24 &3,961 (8) & 0.444 & 14.28 &  0.37 \\ 
			(b) CapNet, \texttt{EM} &  152.44 & 10,078 (4)  & 0.957 & \textbf{12.66} & \textbf{0.31} \\ 
			(c) vanilla CNN, same shape & 24.44 & 1,652 (128) & 0.026  & 14.43 &  0.38\\ \hline 
			(d) vanilla CNN, similar param &  146.88& 2,420 (128)& 0.146 & 12.09 &  0.33\\ \hline
			(e) EncapNet, \texttt{master} &  25.76
			& 1,433 (128)&0.039  & 13.87& 0.31\\ 
			(f) EncapNet, \texttt{master/aide} &  60.68&1,755 (128) & 0.061 & \textbf{11.93}& \textbf{0.25}\\ \hline
		\end{tabular}
	}
	\vspace{-.3cm}
\end{table}
\addtocounter{footnote}{1}
\footnotetext{A single Titan X GPU, which has a 12G memory.}

\section{EncapNet: Neural Network Encapsulation}

\subsection{Approximate routing  with master/aide interaction}\label{sec:approximate-routing--with-master-aide-interaction}

Recall that higher capsules are generated according to the voting co-efficient $c_{ij}$ across all entities (capsules) in the lower layer:
\vspace{-.3cm}
\begin{align}
\bm{s}_j  = \sum_{i=1}^{n_1} c_{ij} \cdot \hat{\bm{v}}_{j | i}
&  =  {c}_{1j}  \hat{  \bm{v}  }_{j| 1}  + \cdots + c_{i j}  \hat{   \bm{v}   }_{j | i}  + \cdots + c_{n_{1}j}  \hat{   \bm{v}   }_{j | n_{1}}, \label{elementwise} \\	
&  =   \underbrace{ c_{i j}   \hat{   \bm{v}   }_{j | i } }_{i = j}+ \sum_{i \neq j} c_{ij} \hat{   \bm{v}   }_{j | i}.
\vspace{-.2cm}
\end{align}
Eqn. (\ref{elementwise}) can be grouped into two parts: 
one is a main mapping that directly receives knowledge from its lower counterpart $i$, whose spatial location is the same as $j$'s; another is a side mapping which sums up  all the remaining lower capsules, whose spatial location is different from $j$'s.
Hence the original unsupervised and short-lived routing process can be approximated in a supervised manner (see Fig. \ref{fig:capsule_block}(c)): 
\begin{align}
\bm{s}_j &  \approx  m_1     \hat{\bm{v}} ^{(1)}_{|{l(\mathcal N_j, k_1)}}  + m_2   \hat{   \bm{v}   }^{(2)}_{| {l}(\overline {\mathcal{N}}_j, k_2)}, \label{approx}
\end{align}
where $\mathcal{N}_j$ is a location set 
along the {channel} dimension that \textit{directly} maps lower capsules (there might be more than one) to higher $j$;
$\overline{\mathcal{N}}_j$ is the complimentary set of $\mathcal{N}_j$ that contains the \textit{remaining} locations along the channel;  $k_{(*)}$ is the spatial kernel size; 
altogether $l(\cdot, \cdot)$ indicates  the location set of all contributing lower capsules to create a higher capsule. Formally, we  define 
$\hat{\bm{v}} ^{(1)}$ and $\hat{\bm{v}} ^{(2)}$
in Eqn. (\ref{approx}) as the  
master and aide activation, respectively, with their co-efficients denoted as $m_1$ and $m_2$.

The \texttt{master} branch looks for the same pattern in two consecutive layers and thus only 
sees a window from its direct lower capsule. 
%
The \texttt{aide} branch, on the other hand, serves as a side unit to replenish information 
from capsules located in other channels. 
The convolution kernels in both branches use the spatial locality: kernels only attend to a small neighborhood of size $k_1 \times k_1$ and $k_2 \times k_2$ on the input capsule $\bm{u}$ to generate the intermediate activation $\hat{\bm{v}} ^{(1)}$ and $\hat{\bm{v}} ^{(2)}$.
The master and aide activations in these two branches are communicated  
by their co-efficients in an interactive manner. Co-efficient $m_{(*)}$ is the output of group convolution; the input source is from both $ \hat{\bm{v}} ^{(1)}$ and $ \hat{\bm{v}} ^{(2)}$, leveraging information encoded in capsules from both the \texttt{master} and \texttt{aide} branches.

\begin{figure}[t]
	\begin{center}
		\includegraphics[width=0.95\textwidth]{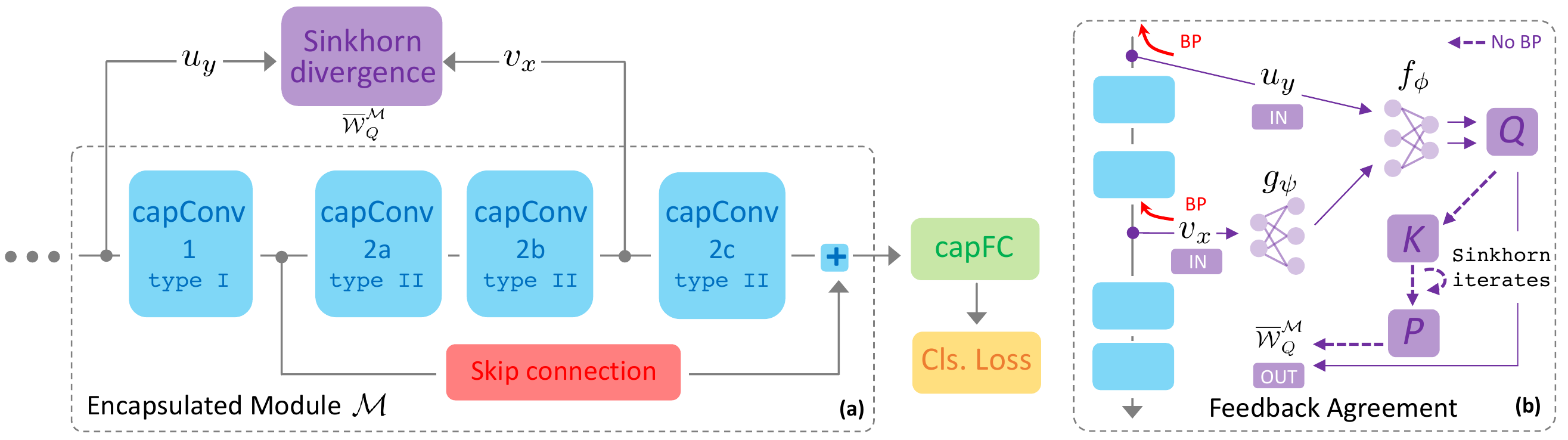}
	\end{center}
	\vspace{-.5cm}
	\caption{
		(a) Connections inside one module of EncapNet, where several \texttt{capConv} layers (type I and II) are cascaded with skip connection and  regularized by the Sinkhorn divergence. This is one type of design and in Section \ref{sec:experiments} we report other variants. (b) Pipeline and gradient flow in the Sinkhorn divergence.
	}
	\label{fig:net-structure}
	\vspace{-.3cm}
\end{figure}

After the interaction as shown in Fig. \ref{fig:capsule_block}(c), we append the batch normalization \cite{bn}, rectified non-linearity unit \cite{relu_icml10} and \texttt{squash} operations at the end. These connectivities are not shown in the figure for brevity. To this end, we have encapsulated one layer of the neural network with each neuron being replaced by a capsule, where interaction among them is achieved by the   \texttt{master/aide} scheme, and denote the whole pipeline as the \texttt{capConv} layer.
An encapsulated module 
is illustrated in Fig. \ref{fig:net-structure}(a), where several \texttt{capConv} layers are cascaded with a skip connection.
There are two types of \texttt{capConv}. Type I is to increase the dimension of capsules across modules and merge spatially-distributed capsules. The kernel size in the \texttt{master} branch is set to be 3 in this type. Type II is to increase the depth of the module for a length of $N$; the  dimension of capsule is unchanged; nor does the number of spatial capsules. The kernel size for the \texttt{master} branch in this type is set to be 1. The \texttt{capFC} block is a per-capsule-dimension operation of the fully-connected layer as in standard neural network. Table \ref{net_structure_ablation} gives an example of the proposed  network, called {EncapNet}.

\textbf{Comparison to CapNet.} 
Compared to the heavy computation of generating a huge number of mappings for each higher capsule in CapNet, our design only requires two mappings in the \texttt{master} and \texttt{aide} branch.
The computational complexity is reduced by a large margin: the kernel size in the transform matrix in the first step is $\frac{n_2}{2}$ times fewer and the routing scheme in the second step is $\frac{S^4}{d_2}$ times fewer ($S$ being the spatial size of feature map). Take the previous setting in Table \ref{net_compare} for instance, 
our design leads to 1024 and 256 times fewer parameters than the original 8,388,608 
and 4,194,304 parameters in these two steps.
To this end, we replace the unsupervised, iterative routing process \cite{capsule,cap_EM} with a supervised, one-pass \texttt{master/aide} scheme. Compared with (a-b) in Table \ref{net_compare}, our proposed method (e-f) has fewer parameters, less runtime, and better performance. It is also observed that the side information from the \texttt{aide} branch is a necessity to replenish the \texttt{master} branch, with baseline error 13.87\% decreasing to 11.93\%  on CIFAR-10, \textit{c.f.} (e) vs. (f) in Table \ref{net_compare}.

\subsection{Network regularization by feedback agreement}\label{sec:feedback-agreement-as-a-regularization}
Motivated by the agreement routing  where higher capsules should be activated if there is a good `agreement' with lower counterparts, 
we include a loss that requires the higher layer to be able to recover the lower layer. 
The influence of such a constraint (loss) is used during training and removed during inference.

To put the intuition aforementioned in math notation, let 
${v}_x = \{\bm{v}_j\}_{j =1}^{n_2} 
$ and ${u}_y = \{\bm{u}_i\}_{i=1}^{n_1}$ be a sample in space $\mathcal{Z} $ and $\mathcal{U} $, respectively, where $x, y$ are sample indices. Consider a  set of observations, \textit{e.g}. capsules at lower layer, $\mathcal{S}_1 = (u_1, \dots, u_y, \dots, u_{\mathcal{B}_1})  \in \mathcal{U}^{\mathcal{B}_1} $, we design a loss 
which enforces  samples  $v$ on space $\mathcal{Z}$ as input (\textit{e.g.}, capsules at higher layer) can be mapped to $u'$ on space $\mathcal{U}$ through a differentiable function $g_{\psi}: \mathcal{Z} \rightarrow \mathcal{U}$, \textit{i.e.}, $u' = g_{\psi} (  {v})$.
The data distribution, denoted as $\mathds{P}_\psi$, for the generated set of samples $\mathcal{S}_2 = (u'_1, \dots, u'_x, \dots, u'_{\mathcal{B}_2})  \in \mathcal{U}^{\mathcal{B}_2}$ should be as  much
\textit{close} as the distribution $\mathds{P}_r$ for $\mathcal{S}_1$.
In summary, our goal is to find $\psi*$  that minimizes a certain loss or distance 
between two distributions
$\mathds{P}_\psi, \mathds{P}_r \in \text{Prob}(\mathcal{U})$\footnote{In some literature, \textit{i.e.}, \cite{wgan,sinkhorn_loss}, it is called the probability measure and commonly denoted as $\mu$ or $\nu$; a coupling is the joint distribution (measure). We use distribution or measure interchangeably in the following context. $\text{Prob}(\mathcal{U})$ is the set of probability distributions over a metric space $\mathcal{U}$.}:
${\arg\min}_{\psi*} \mathcal{L} (\mathds{P}_\psi, \mathds{P}_r)$. 

In this paper, we opt for an optimal transport (OT) metric 
to measure the distance. The OT metric between two joint probability distributions supported on two metric spaces $(\mathcal{U}, \mathcal{U})$ is defined as the solution of the 
linear program  \cite{regularized_OT}:
\begin{equation}
\vspace{-.1cm}
\mathcal{W}_{Q}(  \mathds{P}_\psi, \mathds{P}_r  )= 
\inf_{\gamma \in \Gamma(\mathds{P}_\psi, \mathds{P}_r   )}
\mathds{E} \bigg[ \int_{    
	\mathcal{U} \times \mathcal{U}
} Q(u', u) d \gamma(u', u)  \bigg], \label{ot_loss} \\
\end{equation}
where $\gamma$ is a coupling; $\Gamma$ is the set of couplings that consists of joint distributions over the product space with  marginals $ (\mathds{P}_\psi, \mathds{P}_r )$. Our formulation skips some mathematic notations; details are provided in
\cite{sinkhorn_loss,regularized_OT}. Intuitively, $\gamma(u', u)$ implies how much ``mass'' must be transported from $u'$ to $u$ in order to transform the distribution $\mathds{P}_\psi$ into  $\mathds{P}_r$; $Q$ is the ``ground cost" to move a unit mass from $u'$ to $u$.
As is well known, Eqn. (\ref{ot_loss}) becomes the \textit{p}-Wasserstein distance (or loss, divergence) between probability measures when $\mathcal{U}$ is equipped with a distance $\mathcal{D}_{\mathcal{U}}$ and $Q=\mathcal{D}_{\mathcal{U}}(u', u)^p$, for some exponent $p$.

Note that the expectation $\mathds{E}(\cdot)$ in Eqn. (\ref{ot_loss}) is used for 
mini-batches of size $(\mathcal{B}_1, \mathcal{B}_2)$. In our case, $\mathcal{B}_1$ and $\mathcal{B}_2$ are equal to the training batch size.
Since both input measures are discrete for the indices $x$ and $y$ (capsules in the network),
the coupling $\gamma$ can be treated as a non-negative matrix $P$, namely $\gamma = \sum_{x,y} P_{x, y} \delta(v_x, u_y) \in \text{Prob}(\mathcal{Z} \times \mathcal{U})$, where $\delta$ represents the Dirac unit mass distribution at point $(v, u)\in (\mathcal{Z} \times \mathcal{U})$. Rephrasing the continuous case of Eqn. (\ref{ot_loss}) into a discrete version, we have the desired OT loss:
\begin{equation}
\mathcal{W}_{Q}(  \mathds{P}_\psi, \mathds{P}_r  )  \xleftarrow[]{\text{discrete}} 
\min_{P \in \mathds{R}_{+}^{\mathcal{B}_2 \times \mathcal{B}_1}}
\langle  Q, P\rangle, \label{ot_loss_discrete}
\end{equation}
where $P$ satisfies $P^\mathsf{T} \mathds{1}_{\mathcal{B}_2}  = \mathds{1}_{\mathcal{B}_1}, P\mathds{1}_{\mathcal{B}_1}  = \mathds{1}_{\mathcal{B}_2}$. $ \langle  \cdot, \cdot\rangle$ indicates the Frobenius dot-product for two matrices and $\mathds{1}_m \coloneqq (1/m, \dots, 1/m) \in \mathds{R}_{+}^{m}$.
Now the problem boils down to computing $P$ given some ground cost $Q$.
We adopt the Sinkhorn algorithm \cite{sinkhorn1964}  in an iterative manner, which is promised to have a differentiable loss function \cite{regularized_OT}.
Starting with $b^{(0)} = \mathds{1}_{\mathcal{B}_2}, l \leftarrow 0$, $ \texttt{Sinkhorn iterates} $ read :
\begin{gather}
a^{(l+1)} \coloneqq \frac{ \mathds{1}_{\mathcal{B}_1} }{K^{\mathsf{T}} b^{(l)}},  ~~~
b^{(l+1)} \coloneqq \frac{ \mathds{1}_{\mathcal{B}_2} }{K^{} a^{(l)}},
\end{gather}
where the Gibbs kernel  $K_{x, y}$ is defined as $\exp(-Q_{x,y} / \varepsilon)$; $\varepsilon$ is a control factor. For a given budget of $L$ iterations, we have:
\begin{equation}
P \coloneqq P^{(L)} = \text{diag}(b^{(L)}) \cdot K \cdot \text{diag}(a^{(L)}),
\end{equation}
which serves as a proxy for the OT coupling. 
Equipped with the computation of $P$ and having some form of cost $Q$ in hand, we can minimize the optimal transport divergence  along with other loss in the network.

In practice, we introduce a bias fix to the original OT distance in Eqn. (\ref{ot_loss_discrete}), namely the Sinkhorn divergence 
\cite{sinkhorn_loss}.
Given two sets of samples $v_x, u_y$  and accordingly  distributions $\mathds{P}_\psi, \mathds{P}_r$,  the revision is defined as:
\begin{gather}
\overline{\mathcal{W}}^{\mathcal{M}}_{Q}(\mathds{P}_\psi, \mathds{P}_r) = 2\mathcal{W}_{Q}(  \mathds{P}_\psi, \mathds{P}_r  ) - \mathcal{W}_{Q}(\mathds{P}_\psi, \mathds{P}_\psi) - \mathcal{W}_{Q}(\mathds{P}_r, \mathds{P}_r), \label{sinkhorn_loss}
\end{gather}
where $\mathcal{M}$ is the module index.
By tuning $\varepsilon$ in $K$ from $0$ to $\infty$, the Sinkhorn divergence has the property of 
taking the best of both OT (non-flat geometry) and MMD \cite{MMD} (high-dimensional rigidity) loss, which we find in experiments  improves performance.

The overall workflow to calculate a Sinkhorn divergence\footnote{The term Sinkhorn used in this paper is two-folds: one is to indicate the computation of $P$ via a \texttt{Sinkhorn iterates}; another is to imply the revised OT divergence.} is depicted in Fig. \ref{fig:net-structure}(b). Note that our ultimate goal of applying OT loss is to make  feature learning in the \textit{mainstream} (blue blocks) better aligned across capsules in the network. 
It is added during training and abominated for inference. Therefore the design for Sinkhorn divergence has two principles: light-weighted and capsule-minded.  Sub-networks $g_\psi$ and $f_\phi$ should increase as minimal parameters to the model as possible; the generator should be encapsulated to match the  data structure. 
Note that the Sinkhorn divergence is optimized to minimize loss w.r.t. {both} $\phi,\psi$, 
instead of the practice in \cite{sinkhorn_loss,salimans2018improving,wgan}  via an adversarial manner. 
%

\textbf{Discussions.} (i)
There are alternatives besides the OT metric for  
$\mathcal{L}(\mathds{P}_\psi, \mathds{P}_r)$, 
\textit{e.g.}, the Kullback-Leibler (KL) divergence, which is defined as
$\sum_y \log \frac{\text{d} \mathds{P}_\psi }{\text{d}u'} u_y$ or Jenson-Shannon (JS) divergence. In \cite{wgan}, it is observed that these distances are not sensible when learning distributions supported by low dimensional manifolds on $\mathcal{Z}$. Often the model manifold and the ``true'' distribution's support often have a non-negligible intersection, implying that $KL$ and $JS$ are  non-existent or  infinite  in some cases. 
In comparison, the optimal transport loss 
is continuous and differentiable on $\psi$ under mild assumptions nonetheless.
%
(ii) Our design of feedback agreement unit is not limited to the capsule framework. Its effectiveness on vanilla CNNs is also verified by experimental results in Section \ref{sec:ablative-analysis}.

\textbf{Design choices in OT divergence.}
We use a deconvolutional version of the \texttt{capConv} block as the mapping function $g_\psi$ for reconstructing lower layer neurons from higher layer neurons.
Before feeding into the cost function $Q$, samples from two distributions are passed into a feature extractor $f_\phi$. The extractor is modeled by a vanilla neural network and can be regarded as a dimensionality reduction of $\mathcal{U}$ onto a lower-dimension space. 
There are many options to design the cost function $Q$, such as cosine distance or $l_2$ norm.
Moreover, it is found in experiments that if the gradient flow in the \texttt{Sinkhorn iterates} process is ignored as does in \cite{salimans2018improving}, the result gets slightly better.
Remind that 
$
Q_{x, y} = \mathcal{D}\big(  f_{\phi} ( {u'}_x) , f_{\phi}(  {u}_y)
\big)$ is dependent on $\phi, \psi$ (so does $P, K, a, b$); hence the whole OT unit can be trained in the standard  optimizers (such as Adam \cite{adam_opt}).

\textbf{Overall loss function.} The final loss of EncapNet is a weighted combination from both the Sinkhorn divergence across  modules and the marginal loss \cite{capsule} for capsule in the classification task: $
\mathcal{L}_{\texttt{margin}}(t, \bm{v}) + \lambda  \sum_{\mathcal{M}}  \overline{\mathcal{W}}^{\mathcal{M}}_{Q} $, where $t, \bm{v}$ is the ground truth and  class capsule outputs of the \texttt{capFC} layer, respectively; $\lambda$ is a hyper-parameter to negotiate between these two losses (set to be 10).

\section{Related Work}

\textbf{Capsule network.}
Wang \textit{et al.} \cite{opt_view} formulated the routing process 
as an optimization problem that minimizes the clustering-like loss and a KL regularization term. They proposed a more general way to regularize the objective function, which shares similar spirit as the agglomerative fuzzy \textit{k}-means algorithm \cite{Li2008AgglomerativeFK}.
%
Shahroudnejad \textit{et al.} \cite{cap_relevance_path} explained the capsule network  inherently  constructs a relevance path, by way of dynamic routing in an unsupervised way, to eliminate the need for a backward process. When a group of capsules agree for a parent one, they construct a part-whole relationship which can be considered as a relevance path.
A variant capsule network \cite{spectral_cap} is proposed 
where capsule's activation is obtained based on the eigenvalue of a decomposed voting matrix. 
Such a spectral perspective witnesses a faster convergence and a better result than the EM routing  \cite{cap_EM} on a learning-to-diagnose problem.  

\textbf{Attention vs. routing.}
In \cite{first_attention}, Mnih \textit{et al.} proposed a recurrent module to extract  information by adaptively selecting a sequence of regions and to only attend the selected locations. 
DasNet \cite{sel_attention} allows the network to iteratively focus   attention on  
convolution filters via feedback connections from higher layer to lower ones. The network generates an
observation vector, which is used by a deterministic policy to choose an action,  and accordingly changes the weights of feature maps for better classifying objects.
Vaswani \textit{et al.} \cite{attention_is_all_you_need} formulated a multi-head attention for machine translation task where attention coefficients are  calculated and parameterized by a compatibility function. 
%
%
Attention models aforementioned tries to learn the attended weights from lower neurons to higher ones. The lower activations are weighted by the learned parameters in attention module to generate higher activations. 
However, the agreement routing scheme \cite{capsule,cap_EM} is a top-down solution: higher capsules should be activated if and only if the most similar lower counterparts have a large response. The routing co-efficients is obtained by recursively looking back at lower capsules and updated based on the resemblance. 
Our approximate routing can be deemed as a bottom-up approach which shares similar spirit as attention models.


\section{Experiments}\label{sec:experiments}

The experiments are conducted on 
CIFAR-10/100 \cite{cifar}, 
SVHN \cite{svhn} and a large-scale dataset called ``\texttt{h}-ImageNet''. We construct the fourth one as a subset of the ILSVRC 2012  classification database \cite{imagenet_trans}.
It consists of 200 hard classes whose top-1 accuracy, based on the prediction output of the ResNet-18 \cite{resNet} model is lower than other classes. The ResNet-18 baseline model on \texttt{h}-ImageNet has a 41.83\% top-1 accuracy. The dataset has a collection of 
255725, 17101 
images for training and validation, compared with CIFAR's 50000 for training and 10000 for test. We manually crop the object with some padding for each image (if the bounding box is not provided) since the original image has too much background and might be too large (over 1500 pixels); after the pre-processing, each image size is around 50 to 500, compared with CIFAR's $32$ input.
``\texttt{h}-ImageNet'' is proposed for fast verifying ML algorithms on a  large-scale dataset which shares similar distribution as 
ImageNet. 

\textbf{Implementation details.} 
The general settings are the same across   datasets if not specified afterwards. Initial learning rate is set to 0.0001 and reduced by 90\% with a schedule $[200, 300, 400]$ in epoch unit. Maximum epoch is 600. Adam \cite{adam_opt} is used  with momentum 0.9 and weight decay $5\times10^{-4}$. Batch size is 128.

\subsection{Ablative analysis}\label{sec:ablative-analysis}
In this subsection we analyze the connectivity design in the encapsulated module and the many choices in the OT divergence unit. 
The depth of EncapNet and ResNet are the same  18 layers ($N=3, n=2$) for fair comparison. Their structures are depicted  in Table \ref{net_structure_ablation}.
Remind that the comparison of \texttt{capConv} block with CapNet 
is reported in Table \ref{net_compare} and analyzed in Section \ref{sec:approximate-routing--with-master-aide-interaction}.

\renewcommand\arraystretch{1.2}
\setlength{\tabcolsep}{.8pt}
\begin{table}
	\centering
	\caption{Network architecture of EncapNet and ResNet.
		The compared ResNet variant has the same input and output shape as EncapNet.  
		`$x \rightarrow y$' indicates channel dimension from input to output. \texttt{capConv}($k, s, p$) means the 
		\texttt{master} capsule has a convolution  of kernel size  $k$, stride $s$ and padding $p$.  Similarly for the standard convolution $\texttt{conv}()$ and 
		residual block $\texttt{res}()$. The depth of the EncapNet and ResNet is $2+ \sum_i (N_i+1)$ and $2+ \sum_i 2n_i$, respectively. 
		Connection of OT divergence is omitted for brevity.} \label{net_structure_ablation}
	\vspace{-.2cm}
	\scriptsize	{
		\begin{tabular}{c | c | c | c  | c   | c  }
			\hline
			\multicolumn{2}{ c |  }{module }  & output size & cap dim. & EncapNet\_v1  & ResNet \\ 
			\hline
			\multicolumn{2}{c| }{$\mathcal{M}_0$} &  $32 \times 32$& - &  $3\rightarrow 32$, \texttt{conv}$(3,1,1)$ & $3\rightarrow 32$, \texttt{conv}$(3,1,1)$ \\ 
			\hline
			\multirow{2}{*}{$\mathcal{M}_1$} &  \scriptsize  I & \multirow{2}{*}{ $32 \times 32$  }& $1 \rightarrow 2$ &  $32\rightarrow 32, \texttt{capConv}(3,1,1)$ & $32\rightarrow 64$, \texttt{res}$(3,1,1)$ \\
			& \scriptsize  II &  & $2$ &  $\big[32\rightarrow 32, \texttt{capConv}(1,1,0)  \big] \times N_1$ &$\big[64\rightarrow 64, \texttt{res}(3,1,1)  \big] \times (n_1-1)$ \\ 
			\hline
			\multirow{2}{*}{$\mathcal{M}_2$} & \scriptsize  I & 
			\multirow{2}{*}{$16\times16$}& $2 \rightarrow 4$ & 
			\centering $32\rightarrow 32, \texttt{capConv}(3,2,1)$ &$64\rightarrow 128$, \texttt{res}$(3,2,1)$ \\
			& \scriptsize{II} &  & $4$ &   $\big[32\rightarrow 32, \texttt{capConv}(1,1,0)  \big] \times N_2$ & $\big[128\rightarrow 128, \texttt{res}(3,1,1)  \big] \times (n_2-1)$ \\ 
			\hline
			\multirow{2}{*}{$\mathcal{M}_3$} &  \scriptsize  I & \multirow{2}{*}{$8\times8$} & $4 \rightarrow 8$ &  $32\rightarrow 32, \texttt{capConv}(3,2,1)$ & $128\rightarrow 256$, \texttt{res}$(3,2,1)$ \\
			& \scriptsize  II &  & $8$ &  $\big[32\rightarrow 32, \texttt{capConv}(1,1,0)  \big] \times N_3$ & $\big[256\rightarrow 256, \texttt{res}(3,1,1)  \big] \times (n_3-1)$ \\
			\hline
			\multirow{2}{*}{$\mathcal{M}_4$} &  \scriptsize  I & \multirow{2}{*}{$4\times4$}& $8 \rightarrow 16$ &  $32\rightarrow 32, \texttt{capConv}(3,2,1)$ & $256\rightarrow 512$, \texttt{res}$(3,2,1)$\\
			& \scriptsize  II &  & $16$ &  $\big[32\rightarrow 32, \texttt{capConv}(1,1,0)  \big] \times N_4$ &$\big[512\rightarrow 512, \texttt{res}(3,1,1)  \big] \times (n_4-1)$ \\
			\hline
			\multicolumn{2}{c| }{$\mathcal{M}_5$} & 
			10/100/200
			& 16 & \texttt{capFC} & \texttt{avgPool}, \texttt{FC} \\ \hline
		\end{tabular}
	}
	\vspace{-.3cm}
\end{table}

\textbf{Design in \texttt{capConv} block.} Table \ref{tab:ablative_analysis} (1-4) reports the different incoming sources of the co-efficients $m$ in the \texttt{master} and \texttt{aide} branches. Without using \texttt{aide}, case (1) serves as baseline
where higher capsules are only generated from the master activation. 
Note that the 9.83\% result is already superior than all cases in Table \ref{net_compare}, due to the increase of network depth.
Result show that obtaining $m_x$ from the activation $\hat{\bm{v}}^{(x)}$ in its own branch is better than obtaining from the other activations, \textit{c.f.}, cases (2) and (3).
When the incoming source of co-efficient is from both branches, denoted as ``\texttt{maser/aide}\_v3'' in (4), the pattern information from lower capsules is fully interacted by both \texttt{master} and \texttt{aide} branches; hence we achieve the best result of 7.41\% when compared with cases (2) and (3).
%
Table \ref{tab:ablative_analysis} (5-7) reports the result of adding skip connection based on case (4). It is observed that the skip connection used in both types of the \texttt{capConv} block make the network converge faster and get better performance (5.82\%). Our final candidate model employs an additional OT unit with two Sinkhorn losses imposed on each module. One is the connectivity as shown in Fig. \ref{fig:net-structure}(a) where $v_x$ is half the size of $v_y$; another connectivity  is the same as the skip connection path shown in the figure, where $v_x$ shares the same size with $v_y$; the ``deconvolutional'' generator in this connectivity has a stride of 1. 
It performances better (4.55\%) than using one OT divergence alone (4.58\%). 
%

\begin{figure}
	\begin{minipage}[c]{0.7\textwidth}
		\includegraphics[width=\textwidth]{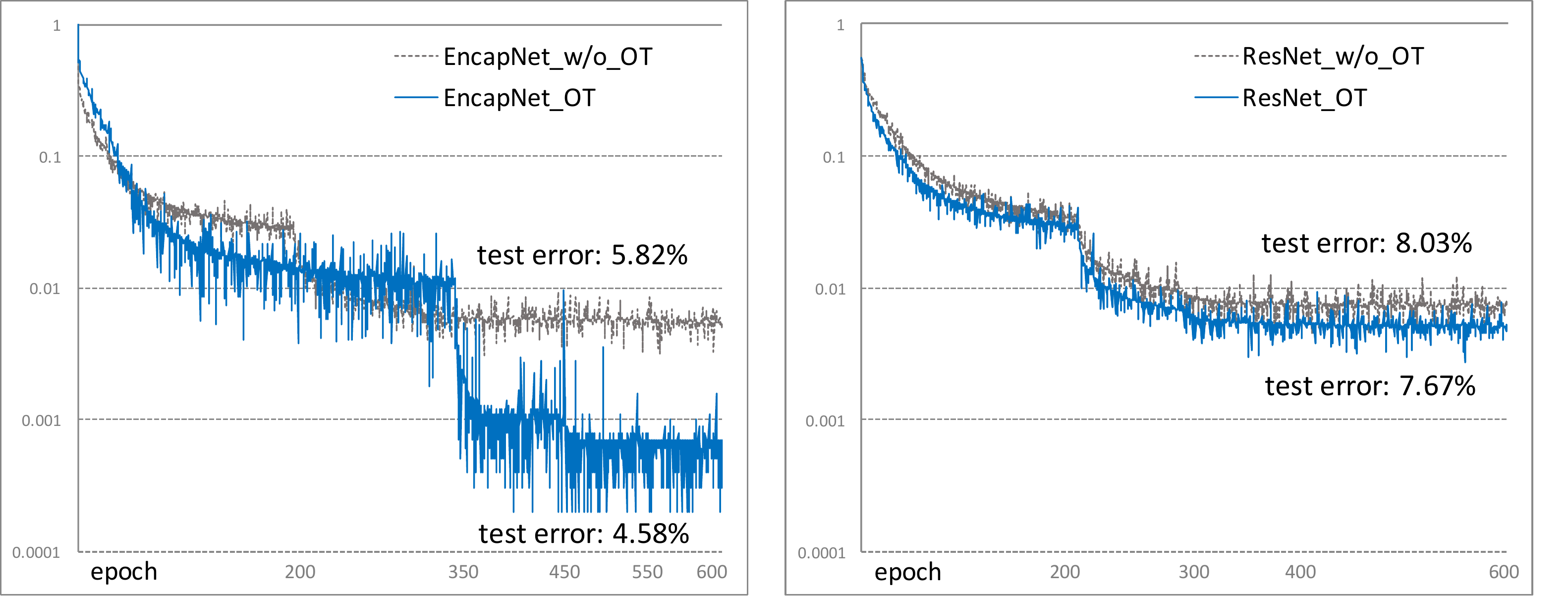}
	\end{minipage}\hfill
	\begin{minipage}[c]{0.28\textwidth}
		\caption{
			Training losses  with embedded optimal transport divergence for  EncapNet and ResNet (*\_OT). One OT unit is adopted as depicted in  Fig. \ref{fig:net-structure}(a) for each module in the network.
		} \label{fig:OT_loss}
	\end{minipage}
	\vspace{-.8cm}
\end{figure}

\textbf{Network regularization design.} Fig. \ref{fig:OT_loss} 
illustrates the training loss curve with and without OT (Sinkhorn) divergence. 
It is found that the performance gain is more evident for EncapNet than ResNet (21\% vs 4\% increase on two networks, respectively).
%
Moreover, we testify the KL divergence option as a distance measurement to substitute the Sinkhorn divergence, shown as case (b) in Table \ref{tab:ablative_analysis}. The error rate  decreases for both model, suggesting that the idea of imposing regularization on the network training is effective; such an add-on is to keep feature patterns better aligned across layers.
The subtlety is that the gain clearly differs when we replace Sinkhorn with KL in EncapNet while these two options barely matter in ResNet.

\renewcommand\arraystretch{1.2}
\setlength{\tabcolsep}{4pt}
\begin{table}[t]
	\caption{ Ablative analysis on (\textbf{left}) the design in the \texttt{capConv} layer and (\textbf{right}) network regularization design. EncapNet and ResNet have the same 18 layers.	``two OTs" indicates each module has two OT divergences coming from different sources. Experiments in series (d-*) are based on case (c) and conducted by removing  or substituting each component in the OT unit while keeping  the rest factors fixed.}\label{tab:ablative_analysis}
	\vspace{-.2cm}
	\scriptsize{
		\centering{
			\resizebox{!}{.14\linewidth}{
				\begin{tabular}{l | c  }
					\hline
					\textbf{\texttt{capConv} Design}  & error  (\%)  \\
					\hline
					(1) \texttt{master} (baseline) & 9.83 \\
					(2) \texttt{maser/aide}\_v1 & 8.05\\
					(3) \texttt{maser/aide}\_v2  & 9.11 \\\hline
					(4) \texttt{maser/aide}\_v3  & 7.41 \\ \hline
					(5) skip\_on\_Type\_I  & 6.81 \\
					(6) skip\_on\_Type\_II  & 6.75\\
					(7) skip\_both & 5.82 \\\hline
					(8) two OTs & \textbf{4.55}\\ \hline
				\end{tabular}
			}
			\resizebox{!}{0.14\linewidth}{
				\begin{tabular}{l | c | c | c}
					\hline
					\textbf{Network Regularization} & \multicolumn{2}{c|}{EncapNet} & ResNet \\
					\hline
					(a) \texttt{capConv} block (baseline) & \multicolumn{2}{c|}{5.82} & 8.03\footnotemark \\ \hline
					(b) KL\_loss  & \multicolumn{2}{c|}{5.31} & 7.72\\ 					\hline
					(c) OT\_loss & \multicolumn{2}{c|}{\textbf{4.58}}& 7.67 \\ \hline
					\multicolumn{1}{l | }{~~~(d1) remove bias fix} & \multicolumn{2}{c|}{4.71} & -  \\
					\multicolumn{1}{l | }{~~~(d2) do BP in $P_L$}& \multicolumn{2}{c|}{4.77}& -\\
					\multicolumn{1}{l | }{~~~(d3) no extractor $f_\psi$}& \multicolumn{2}{c|}{5.79} & - \\
					\multicolumn{1}{l | }{~~~(d4) use vanilla $g_\phi$ } & \multicolumn{2}{c|}{5.01}& -\\
					\multicolumn{1}{l | }{~~~(d5) use $l_2$ in $Q$  }& \multicolumn{2}{c|}{4.90}& -\\	
					\hline
				\end{tabular}
			}
		}
	}
	\vspace{-.3cm}
\end{table}
\addtocounter{footnote}{0}
\footnotetext{ResNet-20 reported in \cite{resNet} has a 8.75\% error rate; some online third party implementation (link anonymised for submission) obtains 6.98\%; we run the 18-layer model in PyTorch with settings stated in the context.}

Furthermore, we  conduct a series of experiments (d-*) to prove the rationale of the Sinkhorn divergence design in Section \ref{sec:feedback-agreement-as-a-regularization}.
Without the bias fix, the result is inferior since it does not leverage both OT and MMD divergences (case d1); if we back-propagate the gradient in the $P_L$ path, the error rate slightly increases; 
the role of feature extractor $f_\psi$ is to down-sample both inputs to the same shape on a 
lower dimension for the subsequent pipeline to process. If we remove this functionality and directly compare the raw inputs $(u,u')$ using cosine distance, the error increases by a large margin to 5.79\%, compared with baseline 5.82\%; if we adopt $l_2$ norm to measure the distance between raw inputs, loss will not converge (not shown in Table \ref{tab:ablative_analysis}). This verifies the urgent necessity of having a feature extractor; if the generator recovering $u'$ from $v$ employs a standard CNN, the performance is inferior (5.01\%) than the capsule version of the generator since data flows in form of capsules in the network; finally if we adopt $l_2$ norm to calculate  $P$ after the feature extractor, the performance degrades as well.

\vspace{-.3cm}

\subsection{Comparison to state-of-the-arts}

As shown in Table \ref{final_compare},
(a) on CIFAR-10/100 and SVHN, we achieve a better performance of 3.10\%,
24.01\% and 1.52 \% compared to previous entires.
The multi-crop test is a key factor to further enhance the result, which is widely used by other methods as well.
(b) on \texttt{h}-ImageNet, 
v1 is the 18-layer structure and has a reasonable top-1 accuracy of 51.77\%. We further increase the depth of EncapNet (known as v2) by stacking more 
\texttt{capConv} blocks, making a depth of 101 to compare with the ResNet-101 model. To ease runtime complexity due to the \texttt{master/aide} intertwined communication, we replace some blocks in the shallow layers with \texttt{master} alone. v3 has a larger input size. Moreover, we have the ultimate version of EncapNet with data augmentation (v3$^{++}$) and obtain an error rate of 40.05\%, compared with the runner-up WRN \cite{wrn}  42.51\%. Training on \texttt{h}-ImageNet roughly takes 2.9 days with 8 GPUs and batch size 256. (c) we have some preliminary results on the ILSVRC-CLS (\texttt{complete}-ImageNet) dataset, which are reported in terms of the top-5 error in Table \ref{final_compare}.

\vspace{-.3cm}
\renewcommand\arraystretch{1.1}
\setlength{\tabcolsep}{1pt}
\begin{table}
	\centering
	\caption{Classification errors (\%) compared to state-of-the-arts. For state-of-the-arts, we show the best results available in their papers.	
		$^{+}$ means mild augmentation while $^{++}$ stands for strong augmentation.
		For \texttt{h}-ImageNet, we train models and report results of other networks  based on the same setting as EncapNet\_v3$^{++}$. 
	} \label{final_compare}
	\vspace{-.2cm}
	\scriptsize{
		\begin{tabular}{r c c c || r | c}
			\hline
			method & CIFAR-10 & CIFAR-100 & SVHN  & \multicolumn{2}{c}{\texttt{h}-ImageNet}  \\
			\hline
			EncapNet  & 4.55
			&  26.77  &  2.01 & EncapNet\_v1 & 48.23 \\
			EncapNet$^{+}$ & 3.13
			& \textbf{24.01}\tiny{(24.85 $\pm$0.11)}& 1.64 & EncapNet\_v2 &  43.15   \\
			EncapNet$^{++}$ & \textbf{3.10} 
			\tiny{(3.56 $\pm$0.12)}& 24.18 & \textbf{1.52}
			\tiny{(1.87 $\pm$0.11)} & EncapNet\_v3 & 42.76    \\
			\cline{1-4}
			GoodInit \cite{iclr16_good_init} & 5.84 & 27.66 &  - & EncapNet\_v3$^{+}$ & 40.18   \\ 
			BayesNet \cite{scalable_bayes} & 6.37 & 27.40 & - &EncapNet\_v3$^{++}$ & \textbf{40.05} \\ \cline{5-6}
			ResNet \cite{resNet} & 6.43 & - & -  &   WRN \cite{wrn} & 42.51     \\
			ELU \cite{iclr16_elu} & 6.55& 24.28 & - &  ResNet-101 \cite{resNet} & 44.13 \\
			Batch NIN \cite{batch_NIN} & 6.75& 28.86 & 1.81 & VGG \cite{vgg} & 55.76    \\
			Rec-CNN \cite{Liang_2015_CVPR} & 7.09 & 31.75 & 1.77 &	  GoogleNet \cite{googlenet} & 60.18   \\ \cline{5-6} 
			Piecewise \cite{piecewise_linear} & 7.51 & 30.83 & -   \\
			DSN \cite{DSN} & 8.22 & 34.57& 1.92 & \multicolumn{2}{c}{\texttt{complete}-ImageNet(top-5) }\\ \cline{5-6}
			NIN \cite{NIN} & 8.80&35.68 &2.35 & EncapNet-18 & 7.51 \\
			dasNet \cite{sel_attention} & 9.22 & 33.78 & - & GoogleNet \cite{googlenet} & 7.89 \\
			Maxout \cite{maxout}      & 9.35  &  38.57   & 2.47 & VGG \cite{vgg} & 8.43\\
			AlexNet \cite{alexnet} & 11.00  &  -   & - & ResNet-101 \cite{resNet} & 6.21\\
			\hline
		\end{tabular}
	}
	\vspace{-1cm}
\end{table}

\section{Conclusions}
In this paper, we  analyze the role of routing-by-agreement to aggregate feature clusters in the capsule network. To lighten  the computational load in the original framework, we devise an approximate routing scheme with master-aide interaction. The proposed alternative is light-weight, supervised and one-time pass during training. The twisted interaction ensures that the approximation can make best out of lower capsules to activate higher capsules. Motivated by the routing process to make capsules better aligned across layers, we send back higher capsules as feedback signal to better supervise the learning across capsules.  Such a network regularization is achieved by minimizing the distance of two distributions using optimal transport divergence during training. This regularization is also found to be effective for vanilla CNNs. 

\smallskip
\noindent\textbf{Acknowledgment.}
We would like to thank Jonathan Hui for the wonderful blog on capsule research, Gabriel Peyré and Yu Liu for helpful discussions in early stage. H. Li and X. Guo are funded by  Hong Kong Ph.D. Fellowship Scheme. 

\bibliographystyle{splncs}
\bibliography{egbib,deep_learning}

\renewcommand\thesection{\Alph{section}}
\renewcommand\thesubsection{\thesection.\roman{subsection}}
\setcounter{section}{0}

\section{More Details and Results in CapNet}
\subsection{Configurations in Section \ref{sec:routing-by-agreement-in-capnet}}\label{sec:supp1}

All the experiments in  Table \ref{net_compare} and Figure \ref{fig:routing_process} employ a shallow network of six layers. 
%
The first four layers across vanilla CNNs, CapNet and EncapNet use the same recipe: 
one standard convolution plus a BN-ReLU follow-up. The stride and padding in convolution is 3 and 1, respectively; these four modules are for increasing the channel number and down-sampling the feature map. After these modules, the activation map is of size $256\times8\times8$.

For {CapNet} (case (a) and (b)), both the fifth and sixth layer are capsule-minded and implemented by group convolution, as shown in Fig. \ref{fig:capsule_block}(b). The transform matrix (kernel in convolution) in the fifth layer is super huge (of size $8, 1, 1, 32\times 16 \times 2048$) since the mapping step involves transforming capsule $i$ to each and every $j$ in the subsequent layer. After routing (dynamic or EM), the output has a shape of $2048\times16$ and is rearranged into the blob shape in a spatial-channel distributed manner ($512 \times 8 \times 8$) for the next layer. 
The budget of routing iteration is set to $R=3$.

There are two subtleties when we implement the routing scheme. For  \texttt{dynamic} routing, we find the loss does not converge if we conduct the softmax operation of co-efficients $b$ along the $j$ dimension (as suggested in 
\cite{capsule});  
\textit{instead we operate softmax  along the $i$ dimension}. For  \texttt{EM} routing, it is observed that \textit{adding a normalization operation} before data is fed into the sigmoid function to generate activation $a_j$ is crucial to make the network stable and faster towards convergence, since the std acquired after the M-step is small. 
Having them through a log function results in most values being into the range, say over 100; this will further generate all 1's after sigmoid. If we add  normalization  to set the range of input data before sigmoid around zero, the output $a_j$ will be diversified in $[0, 1]$.

\smallskip
For the compared {vanilla CNNs} (case (c) and (d)), the fifth module
is a \texttt{conv}-BN-ReLU cascade with average pooling to make spatial size of feature maps to be $1\times1$; the last module is a fully connected layer. The network for case (d), ``similar param'', is achieved by increasing the channel number such that the model size amounts to a similar level as case (a) and (b).

For the compared EncapNet (case (e) and (f)), the fifth layer is a \texttt{capConv} layer with approximate routing and master/aide interaction, which is proposed in the main paper. The sixth layer is a \texttt{capFC} layer with outputs being the same as CapNet's ($10\times 16$).

\subsection{Ablative study on CapNet}

In preliminary experiments, we  conduct some ablative analysis on the routing scheme and different loss choices in CapNet, which is shown in Table \ref{routing_and_loss}. 

\renewcommand\arraystretch{1.1}
\setlength{\tabcolsep}{1pt}
\begin{table}
	\centering
	\caption{The agreement routing and loss choices in CapNet \cite{capsule}. 
		The chosen setting in a certain aspect is marked with gray background. We use dynamic routing and the network structure employs the same 6-layer recipe as stated in Section \ref{sec:supp1}.
	} \label{routing_and_loss}
	\vspace{-.2cm}
	\scriptsize{
		\centering{
			\resizebox{!}{.07\linewidth}{
				\begin{tabular}{c | c  || r | c }
					\hline
					{routing iter.}  & CIFAR-10 & routing co-efficient & CIFAR-10  \\
					\hline
					1 & 14.79& random & 14.35\\
					2 & 14.41 & zero & \cellcolor[gray]{.9}14.28 \\
					3 & \cellcolor[gray]{.9}14.28 & learnable & 14.11\\ \cline{3-4}
					4  & 14.25 \\ \cline{1-2}
				\end{tabular}
			}
			\resizebox{!}{.07\linewidth}{
				\begin{tabular}{r | c  c  c}
					\hline
					loss/optimizer & RMSProp & SGD & Adam \\
					\hline
					spread, fix\_m & 15.61 & 15.38 & 14.97 \\
					spread & 15.02 & 14.87 & 14.50 \\
					margin & 14.30 & 15.18 &  \cellcolor[gray]{.9}14.28\\
					CE\_loss & 14.76 & fail & 15.31\\
					\hline
				\end{tabular}
			}
		}
	}
\end{table}

First, the number of routing iteration. We find the error would get lower if we conduct the routing algorithm via coordinate descent ($R>=2$) than a simple averaging across all lower capsules ($R=1$, implying there is no routing). As the number of iteration goes up, the performance gets slightly better; since it would cost more runtime, we fix $R$ to be 3 considering both efficiency and accuracy.
Second, the initialization of the routing co-efficients. The zero initialization is suggested in 
\cite{capsule};
we also investigate other options. The performance difference between random and learnable initialization of the co-efficients does not vary much. Hence we adopt the original zero manner of initialization.  At last, we investigate different loss choices under various optimizers. In general,  RMSProp and Adam are better in sense of training capsule networks than does SGD. The popular cross entropy loss is inferior than the marginal 
\cite{capsule} 
and spread 
\cite{cap_EM} 
loss.

\section{More Details in EncapNet}

The capsule generator is composed of a \texttt{convTranspose2d} operation with grouping, followed by a BN-ReLU unit; the value of grouping is the dimension of  input capsules. The kernel size in the convolution is 3; the stride could be 2 or 1, depending on whether the size of output activations matches the size of input. The generator appends a squash operation at the end.
The feature extractor (also known as ``critic'')  consists of two consecutive convolutions with each followed by a BN-ReLU sequel. The kernel size, padding and stride of these convolutions are (3,1,2); the number of output channel in the first convolution is one forth of the input's while the second convolution only has one output channel.

The control factor $\epsilon$ to compute the kernel $K$ is set to $0.1$. 
As suggested in 
\cite{learn_w_wass_loss}, 
a stronger regularization (larger $\varepsilon$) 
leads to stable result and faster convergence; hence $L$ can be set small. The length of \texttt{Sinkhorn iterates} is set to $10$.

\section{Comparison to State-of-the-arts}
\vspace{-.1cm}
The mild data augmentation includes the multi-crop test and random cropping; 
and the strong augmentation additionally includes random horizontal flipping, color jittering and longer training. 
For random cropping, a patch of size 32 is randomly cropped out of a resized image of size 34 on CIFAR and SVHN;
a patch of size 128 (224) is randomly cropped out of a resized image of size 156 (256) on \texttt{h}-ImageNet across all versions. The number in parentheses indicates the setting in \texttt{v3} version.
For color jittering, the input color channels are randomly disturbed in brightness, contrast, saturation and hue with $\sigma=0.2$.

The \texttt{h}-ImageNet database\footnote{For detailed statistics,  please refer to the GitHub repo mentioned earlier.} is proposed 
to  train an ImageNet-like dataset in a manageable time. EncapNet\_v3 takes around 2.5 days on 4 GPUs, \textit{c.f.} the full ImageNet database trainig 
\cite{resNet} 
being one week on 8 GPUs. However, the correlation between \texttt{h}-ImageNet and the original database is {not} investigated by time of submission; whether the effectiveness on \texttt{h}-ImageNet \textit{still} holds on ILSVRC for most algorithms will be left as future work.

\end{document}